\documentclass{article}

\PassOptionsToPackage{sort}{natbib}

\usepackage{iclr2024_conference}

\usepackage{times}
\usepackage[utf8]{inputenc} 
\usepackage[T1]{fontenc}    
\usepackage{hyperref}       
\usepackage{url}            
\usepackage{booktabs}       
\usepackage{amsfonts}       
\usepackage{nicefrac}       
\usepackage{microtype}      
\usepackage{graphicx}       
\usepackage{mwe}            
\usepackage{paralist}       
\usepackage{comment}        
\usepackage{soul}           
\usepackage{subcaption}     

\usepackage{tikz}
\usepackage{pgfplots}
\pgfplotsset{compat = 1.18}

\definecolor{bleu}{RGB}{49, 140, 231}

\usepackage{amsmath,amssymb}

\usepackage[british]               {babel}
\usepackage[capitalise, nameinlink]{cleveref}
\usepackage[inline]                {enumitem}

\usepackage{booktabs}       
\usepackage{multicol}       
\usepackage{multirow}       

\usepackage{wrapfig}

\usepackage{color}          
\usepackage{soul}           

\usepackage{siunitx}
\DeclareMathAlphabet{\mathbbold}{U}{bbold}{m}{n}
\newcommand*{\boldone}{\mathbbold{1}}

\hypersetup{
  colorlinks,
  urlcolor  = blue,
  citecolor = blue,
  linkcolor = blue,
}

\title{Differentiable Euler Characteristic \mbox{Transforms} for Shape Classification}
\iclrfinalcopy

\author{%
  Ernst R\"oell$^{1,2}$, Bastian Rieck$^{1,2}$\\[+0.5cm]
  $^1$AIDOS Lab, Institute of AI for Health, Helmholtz Munich\\
  $^2$Technical University of Munich~(TUM)
}

\usepackage{changebar}

\usepackage{xspace}
\usepackage{bbold}
\usepackage{bm}
\newcommand{\modelname}{DECT\xspace}

\begin{document}

\maketitle

\begin{abstract}
  The \emph{Euler Characteristic Transform}~(ECT) is a powerful
  invariant, combining geometrical and topological characteristics
  of shapes and graphs.
  However, the ECT was hitherto unable to learn task-specific
  representations.
  We overcome this issue and develop a novel computational layer that
  enables learning the ECT in an end-to-end fashion.
  Our method, the \emph{Differentiable Euler Characteristic Transform}~(\modelname) 
  is fast and computationally efficient, while exhibiting performance on a par with 
  more complex models in both graph and point cloud classification tasks.
  Moreover, we show that this seemingly simple statistic
  provides the same topological expressivity as more complex topological
  deep learning layers.
\end{abstract}


\section{Introduction} 
\label{sec:Introduction}

Geometrical and topological characteristics play an integral role in
the classification of complex shapes.
Regardless of whether they are represented as point clouds,
meshes~(simplicial complexes), or graphs, the multi-scale perspective
provided by methods from \emph{topological data analysis}~(TDA) can be
applied effectively for classification tasks.
Of particular relevance in this context are the \emph{Persistent Homology
Transform}~(PHT) and the \emph{Euler Characteristic Transform}~(ECT).
Originally introduced by \citet{Turner14a}, recent work proved under
which conditions both transforms are invertible, thus constituting an
injective map~\citep{Ghrist18a, Crawford20a}.
Both transforms are based on the idea of looking at a shape from
multiple directions, and evaluating a multi-scale topological
descriptor for each such direction.
For the PHT, this descriptor is \emph{persistent homology}, a method for
assigning multi-scale topological features to input data, whereas for
the ECT, the descriptor consists of the \emph{Euler characteristic}, an
alternating sum of so-called simplices in a topological space. 
The collection of all these direction--descriptor pairs is then used to
provide a classification or solve an optimisation task.
This approach is mathematically sound, but evaluating \emph{all}
possible directions is infeasible in practice, thus posing a severe
limitation of the applicability of the method.

\paragraph{Our contributions.} We overcome the computational limitations
and present a \emph{differentiable, end-to-end-trainable Euler
Characteristic Transform}. Our method
\begin{inparaenum}[(i)]
  \item is highly scalable,
  \item affords an integration into deep neural networks~(as a layer or
    loss term), and
  \item exhibits advantageous performance in different shape
    classification tasks for various modalities, including graphs, point
    clouds, and meshes.
\end{inparaenum}

\section{Related Work}

We first provide a brief overview of \emph{topological data
analysis}~(TDA) before discussing alternative approaches for shape
classification.
TDA aims to apply tools from algebraic topology to data science
questions; this is typically accomplished by computing algebraic
invariants that characterise the \emph{connectivity} of data.
The flagship algorithm of TDA is \emph{persistent homology}~(PH), which
extracts multi-scale connectivity information about connected
components, loops, and voids from point clouds, graphs, and other data
types~\citep{Barannikov94, Edelsbrunner10}.
It is specifically advantageous because of its robustness
properties~\citep{skraba2020wasserstein}, providing a rigorous approach towards
analysing high-dimensional data.
PH has thus been instrumental for shape analysis and classification,
both with kernel-based methods~\citep{Reininghaus15} and with deep
neural networks~\citep{Hofer17}.
Recent work even showed that despite its seemingly discrete formulation,
PH is differentiable under mild conditions~\citep{moor2020topological,
Carriere21a, Hofer19a, Hofer20}, thus permitting integrations into
standard machine learning workflows.
Of particular relevance for shape analysis is the work by
\citet{Turner14a}, which showed that a transformation based on PH
provides an injective characterisation of shapes.
This transformation, like PH itself, suffers from computational
limitations that preclude its application to large-scale data sets.
As a seemingly less expressive alternative, \citet{Turner14a} thus introduced the
\emph{Euler Characteristic Transform}~(ECT), which is highly efficient
and has proven its utility in subsequent
applications~\citep{amezquita2022measuring, Crawford20a,
marsh2022detecting, nadimpalli2023euler}; see
\citet{munch2023invitation} for an overview.
It turns out that despite its apparent simplicity, the ECT is also
injective, thus theoretically providing an efficient way to characterise
shapes~\citep{Ghrist18a}.
A gainful use in the context of deep learning was not attempted so far,
however, with the ECT and its variants~\citep{jiang2020weighted,
Kirveslahti23a} still being used as \emph{static} feature
descriptors that require domain-specific hyperparameter choices.
\textbf{By contrast, our approach makes the ECT end-to-end trainable, resulting
in an efficient and effective shape descriptor that can be integrated
into deep learning models}.
Subsequently, we demonstrate such integrations both on the level of
\emph{loss terms} as well as on the level of \emph{novel computational
layers}.

In a machine learning context, the choice of model is typically dictated
by the type of data. For \emph{point clouds}, a recent
survey~\citep{Guo21a} outlines a plethora of models for point cloud
analysis tasks like classification, many of them being based on learning
equivariant functions~\citep{zaheer2017deep}.
When additional structure is being present in the form of graphs or
meshes, \emph{graph neural networks}~(GNNs) are typically employed for
classification tasks~\citep{Zhou20a}, with some methods being capable 
to either learn \emph{explicitly} on such higher-order
domains~\citep{Bodnar21a, Bodnar21b, ebli2020simplicial, hajij2020cell,
hacker2020ksimplexvec} or harness their topological
features~\citep{Horn22a, papillon2023architectures}.


\section{Mathematical Background} 
\label{sec:Mathematical background}

Prior to discussing our method and its implementation, we provide
a self-contained description to the \emph{Euler Characteristic
Transform}~(ECT).
The ECT often relies on \emph{simplicial complexes}, the central
building blocks in algebraic topology, which are used extensively for
calculating homology groups and proving a variety of properties
of topological spaces.
While numerous variants of simplicial complexes exist, we will focus on
those that are embedded in $\mathbb{R}^n$.
Generally, simplicial complexes are obtained from a set of
points, to which higher-order elements---\emph{simplices}---such as
lines, triangles, or tetrahedra, are added inductively.
A $d$-simplex~$\sigma$ consists of $d+1$ vertices, denoted by $\sigma
= (v_0, \dots, v_d)$.
A \mbox{$d$-dimensional} simplicial complex $K$ contains simplices up
to dimension~$d$ and is characterised by the properties that
\begin{inparaenum}[(i)]
  \item each face $\tau\subseteq\sigma$ of a simplex~$\sigma$ in $K$ is
    also in $K$, and
  \item the non-empty intersection of two simplices is a face of both.
\end{inparaenum}
Simplicial complexes arise `naturally' when modelling data; for
instance, \emph{3D meshes} are examples of $2$-dimensional simplicial
complexes, with $0$-dimensional simplices being the vertices, the
$1$-dimensional simplices the edges, and $2$-dimensional simplices
the faces; likewise, \emph{geometric graphs}, i.e.\ graphs with
additional node coordinates, can be considered $1$-dimensional
simplicial complexes.

\begin{figure}[tbp]
  \centering
  \subcaptionbox{\label{sfig:StandardFiltration}}{%
    \includegraphics[height=.13\linewidth]{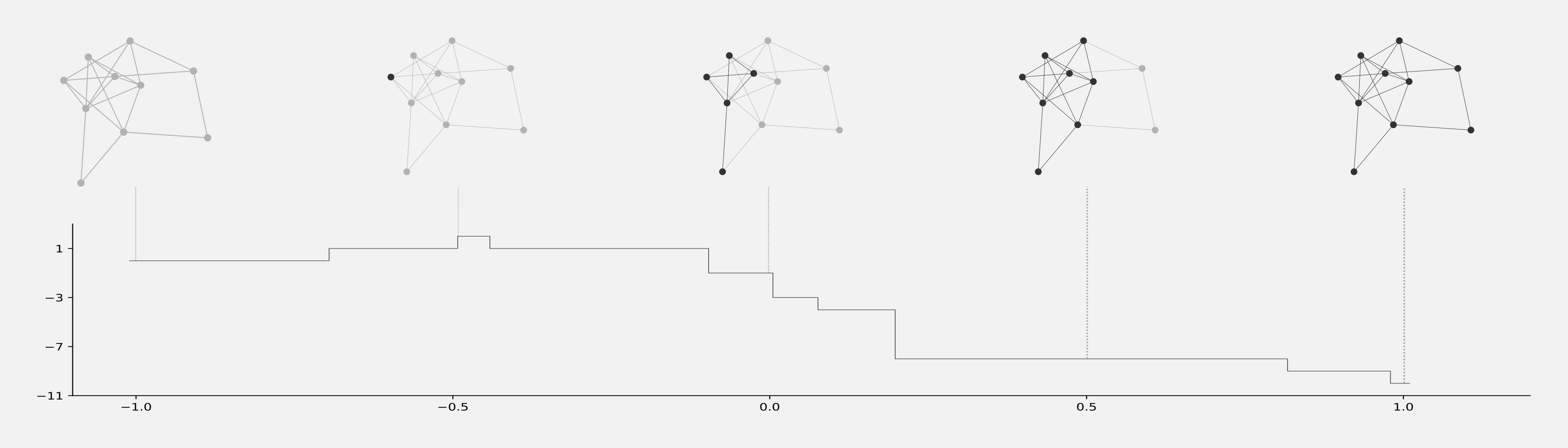}%
  }\quad%
  \subcaptionbox{\label{sfig:InputGraph}}{%
    \includegraphics[height=.13\linewidth]{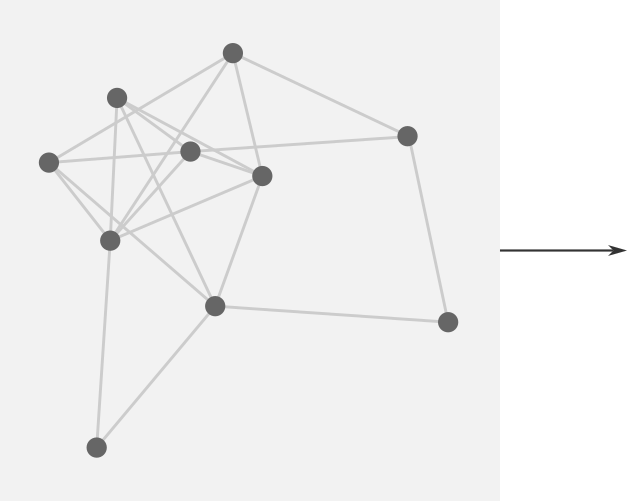}%
  }%
  \subcaptionbox{\label{sfig:ECCs}}{%
    \includegraphics[height=.13\linewidth]{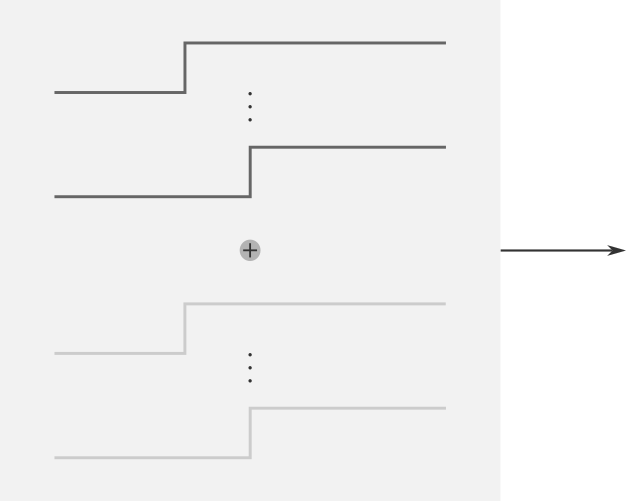}%
  }%
  \subcaptionbox{\label{sfig:FinalCurve}}{%
    \includegraphics[height=.13\linewidth]{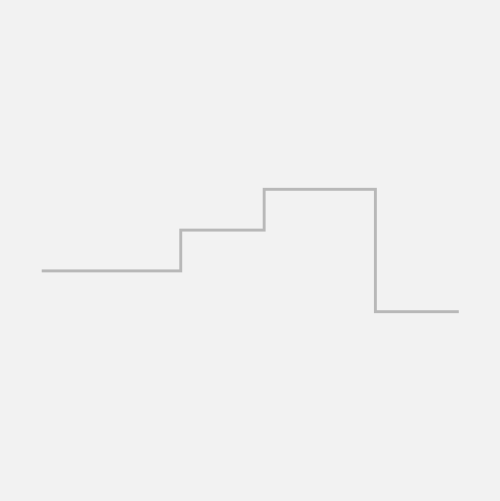}%
  }%
  \caption{%
    The standard algorithm to compute the ECC for a graph, depicted in \subref{sfig:StandardFiltration}, 
    calculates the vertex filtration heights and sorts them in ascending order. 
    One then loops over each set of predefined height values and keeps a running 
    total of the Euler Characteristic as the number of vertices minus edges with 
    height value less than the current height.   
    Our approach differs in that we calculate the ECC of a graph
    \subref{sfig:InputGraph} for each vertex and edge \emph{separately}
    \subref{sfig:ECCs}.
    The sum of the curves is computed for the edges and vertices and the
    total is subtracted to yield the final ECC \subref{sfig:FinalCurve}. 
    \textbf{The advantage is a fully parallel computation, making our
    formulation amenable to hardware accelerations}. 
  }%
  \label{fig:ECTParallel}
\end{figure}
 

\paragraph{Euler characteristic.}
%
Various geometrical or topological properties for characterising
simplicial complexes exist.
One such property is the \emph{Euler characteristic}, defined as the 
alternating sum of the number of simplices in each dimension.
For a simplicial complex $K$, we define the Euler Characteristic $\chi$ as 
\begin{equation}
  \chi(K) = \sum_{n=0} (-1)^n |K^n|,
\end{equation}
where $|K^n|$ denotes the cardinality of set of $n$-simplices.
The Euler characteristic is \emph{invariant} under homeomorphisms and can
be related to other properties of~$K$. For instance, $\chi(K)$ can be
equivalently written as the alternating sum of the \emph{Betti numbers}
of~$K$. Moreover, the Euler characteristic can be defined for other
combinatorial complexes and structures~\citep{Robins02}.

\paragraph{Filtrations.} 
%
The Euler characteristic is limited in the sense that it only
characterises a simplicial complex~$K$ at a single scale.
A multi-scale perspective of this statistic is known to enhance the expressivity 
of the resulting representations.
Specifically, given a simplicial complex~$K$ and a function
$f\colon\mathbb{R}^n \to \mathbb{R}$, we obtain a multi-scale view
on~$K$ by considering the function $\tilde{f}$ as the restriction of~$f$
to the $0$-simplices of~$K$, and defining $\tilde{f}(\sigma) :=
\max_{\tau \subset \sigma} \tilde{f}(\tau)$ for higher-dimensional
simplices.
With this definition, $\tilde{f}^{-1}((-\infty,r])$ is either empty or
a non-empty simplicial subcomplex of~$K$; moreover, for $r_1 \leq r_2$,
we have $\tilde{f}^{-1}((-\infty,r_1]) \subseteq f^{-1}((-\infty,r_2])$.
A function~$\tilde{f}$ with such properties is known as
a \emph{filter function}, and it induces a \emph{filtration} of~$K$ into
a sequence of nested subcomplexes, i.e.\
\begin{equation}
  \emptyset = K_0 \subseteq K_1 \dots \subseteq K_{m-1} \subseteq K_m = K.
\end{equation}
Since the filter function was extended to~$K$ by calculating the
maximum, this is also known as the \emph{sublevel set filtration of~$K$
via~$f$}.\footnote{%
  There is also the related concept of a \emph{superlevel set
  filtration}, proceeding in the opposite direction.
  The two filtrations are equivalent in the sense that they have the
  same expressive power.
}
Filter functions~(and their induced filtrations) can be
learned~\citep{Hofer20, Horn22a}, or they can be defined based on
existing geometrical-topological properties of the input data.
Calculating invariants alongside this filtration results in substantial
improvements of the predictive power of methods.
For instance, calculating homology groups of each $K_i$ leads to
\emph{persistent homology}, a shape descriptor for point clouds.
However, persistent homology does not exhibit favourable scalability
properties, making it hard to gainfully use in practice.


\section{Methods} 
\label{sec:Methods}

\begin{figure}[tbp]
  \centering
  \subcaptionbox{\label{sfig:Filter Graph}}{%
    \includegraphics[height=.25\linewidth]{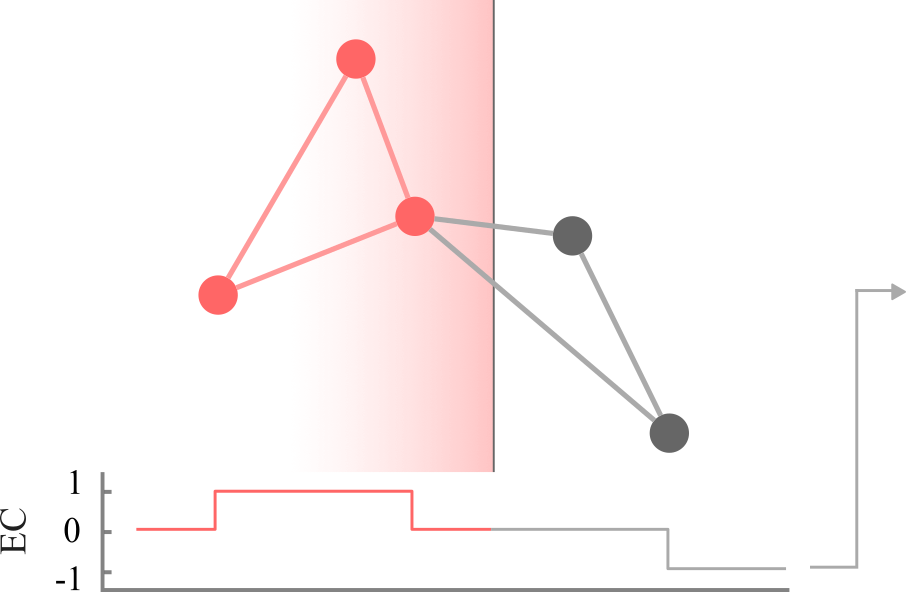}%
  }\quad%
  \subcaptionbox{\label{sfig:Stack Curves}}{%
    \includegraphics[height=.25\linewidth]{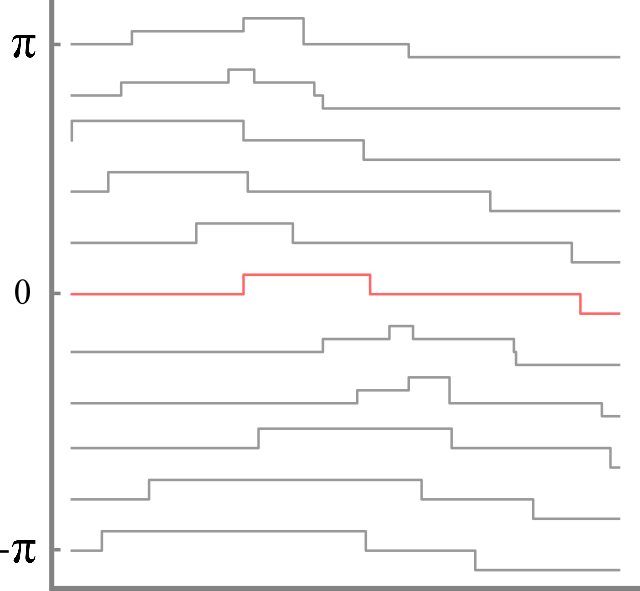}%
  }
  \quad%
  \subcaptionbox{\label{sfig:Final Result}}{%
    \includegraphics[height=.25\linewidth]{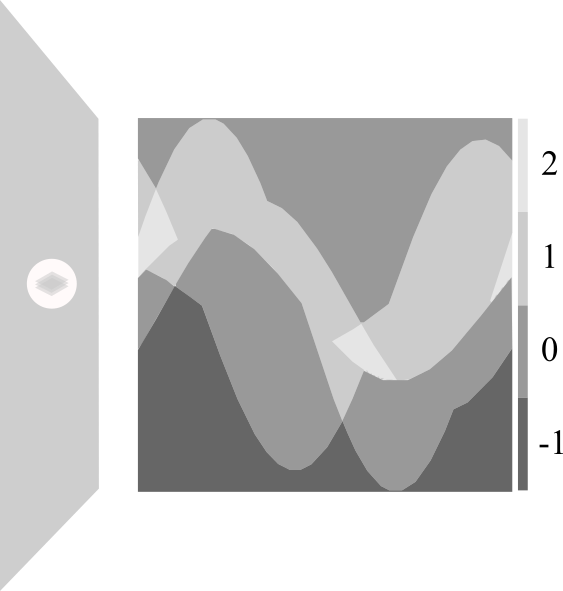}%
  }
  \caption{%
    Overview of the computation of the Euler Characteristic Transform.
    \subref{sfig:Filter Graph}:
    Given a graph and a direction, we filter it with a hyperplane~(here:
    from left to right).
    The nodes and edges of the induced graph are highlighted in red, and the 
    Euler Characteristic Curve of the graph in this direction is 
    displayed below. 
    By the maximum extension principle, edges are added once both target and 
    source node are below the hyperplane.
    \subref{sfig:Stack Curves}:
    We compute the ECC in multiple directions. The curve in \subref{sfig:Filter Graph} 
    is highlighted in red.
    On the vertical axis, we parametrise the direction and on the
    horizontal axis the height.
    \subref{sfig:Final Result}: The ECCs are stacked to form an image,
    where the intensity denotes the Euler Characteristic. This serves
    as the input for machine learning algorithms.
  }%
  \label{fig:overview}
\end{figure}

With the \emph{Euler characteristic} being insufficiently expressive and
\emph{persistent homology} being infeasible to calculate for large data
sets, the \emph{Euler Characteristic Transform}~(ECT), created by
\citet{Turner14a}, aims to strike a balance between the two.
Given a simplicial complex~$K$ and a filter function~$f$,\footnote{%
  For notational simplicity, we drop the tilde from the function
  definition and assume that~$f$ constitutes a valid filter function as
  defined above.
}
the central idea of the ECT is to compute the Euler characteristic
alongside a filtration, thus obtaining a \emph{curve} that serves to
characterise a shape~(see \cref{fig:ECTParallel}).
If the vertices of~$K$ have coordinates in $\mathbb{R}^n$, the ECT is
typically calculated based on a parametric filter function of the form
\begin{equation}
  \begin{aligned}
    f\colon S^{n-1} \times \mathbb{R}^n & \to \mathbb{R}\\
    (\xi, x) &\mapsto \langle x, \xi \rangle,
  \end{aligned}
\end{equation}
where $\xi$ is a \emph{direction}~(viewed as a point on a sphere of appropriate
dimensionality), and $\langle \cdot,\cdot \rangle$ denotes the standard
inner product.
For a fixed $\xi$, we write $f_\xi := f(\xi,\cdot)$.
Given a \emph{height} $h \in \mathbb{R}$, we obtain
a filtration of~$K$ by computing the preimage $f^{-1}_\xi((-\infty,h])$.
The ECT is then defined as:
\begin{equation}
  \begin{aligned}
    \mathrm{ECT}\colon S^{n-1}\times \mathbb{R} &\to \mathbb{Z}\\
    (\xi, h) &\mapsto \chi\left(f^{-1}_\xi \big((-\infty,h]\big)\right),
  \end{aligned}
  \label{eq:ECT}
\end{equation}
If $\xi$ is fixed, we also refer to the resulting curve---which is only
defined for a single direction---as the \emph{Euler Characteristic
Curve}~(ECC).
The ECT is thus the collection of ECCs calculated from different
directions~(see \cref{fig:overview}).
Somewhat surprisingly, it turns out that, given a sufficiently large
number of directions~$\xi$~\citep{curry2022many}, the ECT is
\emph{injective}, i.e.\ it preserves equality~\citep{Turner14a,
Ghrist18a}.

While the injectivity makes the ECT an advantageous shape descriptor, it
is currently only used as a static feature descriptor in machine
learning applications, relying on a set of pre-defined directions~$\xi$, such
as directions chosen on a grid.
We adopt a novel perspective here, showing how to turn the ECT into
a differentiable shape descriptor that affords the integration into deep
neural networks, either as a layer or as a loss term.
Our \textbf{key idea} that permits the ECT to be used in a differentiable
setting is the observation that it can be written as 
\begin{equation}
  \begin{aligned}
    \mathrm{ECT}\colon S^{n-1}\times\mathbb{R} &\to \mathbb{Z}\\
    (\xi,h) &\mapsto \sum_{k}^{\dim K} (-1)^{k} \sum_{\sigma_k} \boldone_{[-\infty,h_{\xi}(\sigma_k))}(h),
  \end{aligned}
  \label{eq:ECT indicator functions}
\end{equation}
where $\sigma_k$ is a $k$-simplex and $h_{\xi}(\sigma_k)$ is the maximum of the 
heights in the direction $\xi$ of the vertices that span $\sigma_k$.
\cref{eq:ECT indicator functions} rewrites the ECT as an alternating sum of
\emph{indicator functions}.
To see that this is an equivalent definition, it suffices to note that for the 
$0$-dimensional simplices we indeed get a sum of indicator functions, as the 
ECT counts how many points are below or above a given hyperplane.
This value is also unique, and once a point is included, it will remain
included.
For the higher-dimensional simplices a similar argument holds.
The value of the filter function of a higher-dimensional simplex is
fully determined its vertices, and once such a simplex is included by
the increasing filter function, it will remain included. 
This justifies writing the ECT as a sum of indicator functions. 

\paragraph{Differentiability.}
%
A large obstacle towards the development of \emph{topological machine
learning} algorithms involves the integration into deep neural networks,
with most existing works treating topological information as mere static
features.
We want our formulation of the ECT to be differentiable with respect to
both the \emph{directions}~$\xi$ as well as the \emph{coordinates}
themselves.
However, the indicator function used in \cref{eq:ECT indicator
functions} constitutes an obstacle to differentiability.
To overcome this, we replace the indicator function with a \emph{sigmoid
function}, thus obtaining a smooth approximation to the ECT.
Notably, this approximation affords gradient calculations.
Using a hyperparameter~$\lambda$, we can control the tightness of the
approximation, leading to
\begin{equation}
  \begin{aligned}
    \mathrm{ECT}\colon S^{n-1}\times\mathbb{R} &\to \mathbb{Z}\\
    (\xi,h) &\mapsto \sum_{k}^{\dim K} (-1)^{k} \sum_{\sigma_k} S\left(\lambda \left( h - h_{\xi}(\sigma_k)  \right)\right),
  \end{aligned}
  \label{eq:ECT sigmoid function}
\end{equation}
where $S(\cdot)$ denotes the sigmoid function.
Each of the summands is differentiable with respect to $\xi$,  
$x_{\sigma_k}$ (the vertex coordinates that span $\sigma_k$), and $h$, 
thus resulting in a highly-flexible framework for the ECT. 
We refer to this variant of the ECT as the
\emph{Differentiable Euler Characteristic Transform}~(\modelname).
Our novel formulation can be used in different contexts, which we will
subsequently analyse in the experimental section.
First, \cref{eq:ECT sigmoid function} affords a formulation as a shape
descriptor layer, thus enabling representation learning on different
domains and making a model `topology-aware.'
Second, since \cref{eq:ECT sigmoid function} is differentiable with
respect to the input coordinates, we can use it to create \emph{loss
terms} and, more generally, optimise point clouds to satisfy certain
topological constraints.
In contrast to existing works that describe topology-based
losses~\citep{Gabrielsson20a, trofimov2023learning, vandaele2022topologically,
moor2020topological}, our formulation is highly scalable without
requiring subsampling strategies or any form of
discretisation in terms of $\xi$~\citep{nadimpalli2023euler}.

\paragraph{Integration into deep neural networks.}
%
Next to being differentiable, our formulation also lends itself to
a better integration into deep neural networks.
Traditionally, methods that employ ECTs for classification concatenate
the ECCs for different directions into a \emph{single} vector, which is
subsequently used as the input for standard classification algorithms,
after having been subjected to dimensionality
reduction~\citep{amezquita2022measuring, jiang2020weighted}.
However, we find that discarding the directionality information like
this results in a loss of crucial information.
Moreover, the concatenation of the ECCs requires the dimensionality 
reduction techniques to be block permutation invariant, as reordering the 
ECCs should \emph{not} change the output of the classification. 
This aspect is ignored in practice, thus losing the interpretability of
the resulting representation.
By contrast, we aim to make the integration of our variant of the ECT
\emph{invariant} with respect to reordering individual curves.
Instead of using a static dimensionality reduction method, we use an MLP
to obtain a learnable embedding of individual Euler Characteristic
Curves into a high-dimensional space.
This embedding is permutation-equivariant by definition.
To obtain a permutation-invariant representation, we use a \emph{pooling
layer}, similar to the \emph{deep sets}
architecture~\citep{zaheer2017deep}.
Finally, we use a simple classification network based on another MLP.
We note that most topological machine learning architectures require
a simplicial complex with additional connectivity information to work. 
This usually requires additional hyperparameters or, in the case of
persistent homology, a sequence of simplicial complexes encoding the
data at multiple scales.
Other deep learning methods, such as deep sets, require a restriction on
the number of points in each sample in the dataset.
By contrast, our method can \emph{directly} work with point clouds,
exhibiting no restrictions in terms of the number of points in each
object nor any restrictions concerning the type of sample connectivity
information.
Hence, \modelname can handle data consisting of a mixture of
point clouds, graphs, or meshes \emph{simultaneously}.

\paragraph{Computational efficiency and implementation.}
%
While there are already efficient algorithms for the computation of the
ECT for certain data modalities, like image and voxel data~\citep{wang2022gpu},
our method constitutes the first description of a differentiable variant
of the ECT in general machine learning settings.
Our method is applicable to point clouds, graphs, and meshes.
To show its computational efficiency, we provide a brief overview on how
to implement \cref{eq:ECT sigmoid function} in practice:
\begin{compactenum}
  \item We first calculate the inner product of all coordinates with
    each of the directions, i.e.\ with each of the coordinates from
    $S^{n-1}$.
  \item We extend these inner products to a valid filter function by
    calculating a \emph{sublevel set filtration}.
  \item We translate all indicator functions by the respective
    filtration value and sample them on a regular grid in the range of
    the sigmoid function, i.e.\ in $[-1,1]$.
    This is equivalent to evaluating $\boldone_{[h_{\xi}(\sigma_k),1]}$ on the
    interval $[-1,1]$.
  \item Finally, we add all the indicator functions, weighted by
    $\pm 1$ depending on the dimension, to obtain the ECT.
\end{compactenum}
All these computations can be \emph{vectorised} and executed in
parallel, making our reformulation highly scalable and benefit from GPU
parallelism.\footnote{%
  Our code is publicly available under \url{https://github.com/aidos-lab/DECT}.
}


\section{Experiments} 
\label{sec:Experiments}

Having described a novel, differentiable variant of the \emph{Euler
Characteristic Transform}~(ECT), we conduct a comprehensive suite of
experiments to explore and assess its properties.
First and foremost, building on the intuition of the ECT being
a universal shape descriptor, we are interested in understanding how
well ECT-based models perform across \emph{different} types of data, 
such as point clouds, graphs, and meshes.
Moreover, while recent work has proven theoretical bounds on the number
of directions required to unique classify a shape~(i.e.\ the number of
directions required to guarantee injectivity) via the
ECT~\citep{curry2022many}, we strive to provide practical insights into
how well classification accuracy depends on the number of directions
used to calculate the ECT.
Finally, we also show how to use the ECT to \emph{transform} point
clouds, taking on the role of additional optimisation objectives that
permit us to adjust point clouds based on a target ECT.

\paragraph*{Preprocessing and experimental setup.}
%
We preprocess all data sets so that their vertex
coordinates have at most unit norm. We also centre vertex coordinates at
the origin.
This scale normalisation simplifies the calculating of ECTs and enables
us to use simpler implementations.
Moreover, given the different cardinalities and modalities of the data, we
slightly adjust our training procedures accordingly.
We split data sets following an $80\%/20\%$ train/test split, reserving
another $20\%$ of the training data for validation.
For the graph classification, we set the maximum number of epochs
to~$100$.
We use the ADAM optimiser with a starting learning rate
of $0.001$.
As a loss term, we either use \emph{categorical cross entropy} for
classification or the \emph{mean squared error}~(MSE) for optimising
point clouds and directions.

\paragraph*{Architectures.}
%
We showcase the flexibility of \modelname by integrating it into
different architectures.
Our architectures are kept purposefully \emph{simple} and do not make
use of concepts like attention, batch normalisation, or weight decay. 
For the synthetic data sets, we add \modelname as the first layer of an
MLP with~$3$ hidden layers. 
For graph classification tasks, we also use \modelname as the first
layer, followed by two convolutional layers, and an MLP with~$3$
hidden layers for classification.
By default, we use $16$ different directions for the calculation of the
ECT and discretise each curve into $16$ steps.
This results in a $16\times 16$ `image' for each input data set. 
When using convolutional layers, our first convolutional layer has $8$
channels, followed by a layer with $16$ channels, which is subsequently
followed by a pooling layer.
Our \emph{classification network} is an MLP with $25$ hidden units
per layer and $3$ layers in total.
Since we represent each graph as a $16\times 16$ image the number of parameters 
is always constant in our model, ignoring the variation in the
dimension of the nodes across the different datasets.
We find that this makes the model highly scalable.

\subsection{Learning directions for classification}

\begin{wraptable}[10]{l}{0.25\linewidth}
    \centering
    \vspace{-\baselineskip}
    \caption{
      By learning directions, \modelname increases accuracy and decreases variance.
    }
    \label{tab:ablations}
    \begin{tabular}{@{}ll@{}}
        \toprule
        \multicolumn{2}{c}{ECT + MLP}\\
        \midrule
        Fixed     & $77.61 \pm 7.98$\\
        Learnable & $81.29 \pm 3.39$\\
        \bottomrule
    \end{tabular}
\end{wraptable}

As a motivating example, we study how learning directions affects the 
classification abilities of \modelname.
We use the MNIST dataset with each non-zero pixel viewed as a point in a point cloud.        
For this experiment we use the MLP model and limit the number of directions to $2$ 
instead of $16$, chosen uniformly on the unit circle.
In the first experiment we keep the directions fixed and in the second we 
allow \modelname to learn the optimal directions.
Both models are trained for $10$ epochs and the experiment is repeated 
$10$ times. 
\cref{tab:ablations} depicts the results and it shows that learning directions 
allows the model to improve the classification accuracy under sparse 
conditions.
For the ECT to be injective in 2D, the minimum number of directions needed is 
$3$ 
when the vertex coordinates of the dataset are known. 
When the vertex coordinates are not known in advance, the minimum number of 
directions depends on the cardinality of the point cloud.
The experiment shows that expressivity is not limited when the number of directions is much lower 
than both the theoretical number of directions needed for injectivity and 
the cardinality of the point cloud.
%

\subsection{Optimising Euler Characteristic Transforms}

\begin{figure}[tbp]
  \centering
  \subcaptionbox{Learning directions\label{sfig:ECT learn directions}}{%
    \includegraphics[width=0.40\linewidth]{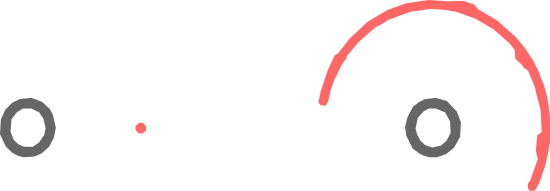}%
  }\quad%
  \subcaptionbox{Learning coordinates\label{sfig:ECT learn point cloud}}{%
    \includegraphics[width=0.40\linewidth]{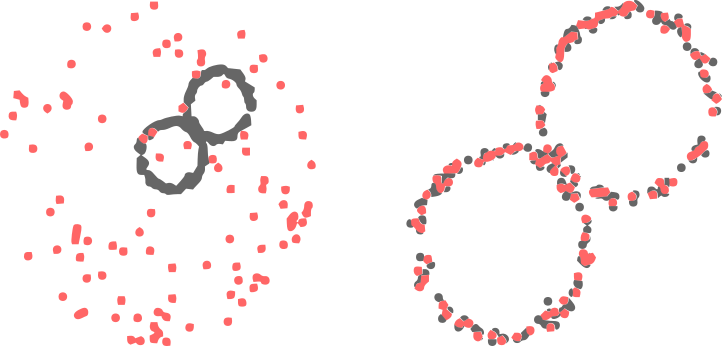}%
  }%
  \caption{%
    \subref{sfig:ECT learn directions}:
    We sample a noisy point cloud from a circle~(grey).
    Red dots show the directions, i.e.\ \emph{angles}, used for the
    ECT~(left: initial, right: after training).
    Our method \modelname spreads directions properly over the unit
    circle, resulting in a perfect matching of the ground truth.
    \subref{sfig:ECT learn point cloud}:
    \modelname also permits us to optimise existing point clouds to
    match a target ECT in an end-to-end differentiable fashion.
    Using two point clouds~(grey: target; red: input data), we train
    \modelname with an MSE loss between the learned ECT and the target
    ECT.
    Starting from a randomly-initialised point cloud~(left), point
    coordinates are optimised to match the desired shape~(right).
    Notably, this optimisation \emph{only} involves the ECT,
    demonstrating its capabilities as a universal shape descriptor.
  }%
  \label{fig:ECT learning example}
\end{figure}

Our method also lends itself to be used in an optimisation setting.
In contrast to prior work~\citep{Gabrielsson20a, moor2020topological, Carriere21a}, 
representations learned by \modelname
permit better \emph{interpretability} since one can analyse which
directions are used for the classification.
The learned directions provide valuable insights into the complexity of
the data, highlighting symmetries.

\paragraph{Learning and visualising directions.}
%
We fix a noisy point cloud sampled from a circle, computing the full
`ground truth' ECT with respect to a set of directions sampled uniformly from~$S^1$.
We then initialise our method \modelname with a set of directions set to
a random point on the unit circle.
Using an MSE loss between the ground truth ECT and the ECT used in our
model, we may \emph{learn} appropriate directions.
\cref{sfig:ECT learn directions} shows the results of the training
process.
We observe two phenomena: first, due to the symmetry of the ECT, it
suffices to only cover half the unit circle in terms of directions;
indeed, each vertical slice of the ECT yields an ECC, which can also be
obtained by rotation.
%
%
The same phenomenon occurs, \emph{mutatis mutandis}, when directions are
initialised on the other side of the circle: the axis of symmetry runs
exactly through the direction closest and furthest from the point cloud,
corresponding to the `maximum' and `minimum' observed in the
sinusoidal wave pattern that is apparent in the ground truth ECT.
We observe that the learned directions are not precisely
situated on the unit circle but close to it.
This is due to \modelname not using a spherical constraint, i.e.\
learned directions are just considered to be points in $\mathbb{R}^2$ as
opposed to being angles.\footnote{%
  We added spherical constraints for all other classification scenarios
  unless explicitly mentioned otherwise.
}
However, the optimisation process still forces the directions to
converge to the unit circle, underpinning the fact that \modelname
in fact learns the ECT of objects even if given more
degrees of freedom than strictly required. 

\paragraph*{Optimising point clouds.} 
%
Complementing the previous experiment on ECT-based optimisation, we also
show how to use \modelname to \emph{optimise} point cloud coordinates
to match a desired geometrical-topological descriptor.
This type of optimisation can also be seen as an additional
\emph{regularisation} based on topological
constraints.
In contrast to existing work~\citep{vandaele2022topologically,
moor2020topological, trofimov2023learning}, our method is
computationally highly efficient and does not necessitate the existence of
additional simplicial complexes.https://people.math.ethz.ch/~skalisnik/
To use \modelname as an optimisation objective,
we normalise all ECTs, thus ensuring that they operate on the same order
of magnitude for an MSE loss.\footnote{
  This is tantamount to making \modelname scale-invariant. We plan on
  investigating additional invariance and equivariance properties in
  future work.
}
Being differentiable, \modelname permits us to adjust the coordinate
positions of the source point cloud as a function of the MSE loss,
computed between the ECT of the model and the ECT of the target point
cloud.
\cref{sfig:ECT learn point cloud} illustrates that \modelname is capable
of adjusting coordinates appropriately.
Notably, this also permits us to train with different sample sizes, thus
creating \emph{sparse approximations} to target point clouds.
We leave the approximation of structured objects, such as graphs or
simplicial complexes, for future work; the higher complexity of such
domains necessitates constructions of auxiliary complexes, which need to
be separately studied in terms of differentiability.

\subsection{Classifying Geometric Graphs}

\begin{table}[tbp]
  \centering
  \caption{
   Comparing \modelname with other methods on the
    \texttt{MNIST-Superpixel} data set.
    We report overall accuracy~($\uparrow$) and runtime per
    epoch~($\downarrow$), highlighting the fact that even on commodity
    hardware, our method is an order of magnitude faster than the
    fastest GNN methods, yielding a favourable trade-off between
    performance, scalability, and accuracy.
    Accuracy can be further improved by using a complex constructed from
    the input images. At the cost of increased runtime for processing
    faces in the complex, our ECT+MLP method is on a par with more
    complex graph neural networks.
    Accuracy values and runtimes of all comparison partners are
    taken from \citet{Dwivedi23a}.
  }
  \label{tab:MNIST}
  \sisetup{
     detect-all              = true,
     table-format            = 2.2(2),
     separate-uncertainty    = true,
     retain-zero-uncertainty = true,
     table-text-alignment    = left,
  }%
  \let\b\bfseries%
  \small%
  \begin{tabular}{lSSSS[table-format=3.2]}
  \toprule
  Method                             & {Accuracy}   & {Epoch runtime~(\si{\second})}\\
  \midrule
  GAT~\citep{velickovic2018graph}    & 95.54\pm0.21 &  42.26\\
  GCN~\citep{kipf2017semisupervised} & 90.71\pm0.22 &  83.41\\
  GIN~\citep{xu2018how}              & 96.49\pm0.14 &  39.22\\
  GraphSage~\citep{Hamilton17a}      & 97.31\pm0.10 & 113.12\\
  MLP                                & 95.34\pm0.14 &  22.74\\
  \midrule
  ECT+CNN (ours)                     & 93.00\pm0.80 &   4.50\\
  ECT+MLP (ours)                     & 97.20\pm0.10 &  10.80\\\bottomrule
  \end{tabular}
\end{table}

Moving from point clouds to graphs, we first study the performance of
our method on the \texttt{MNIST-Superpixel} data
set~\citep{Dwivedi23a}.
This data set, being constructed from image data, has a strong
underlying geometric component, which we hypothesise our model should be
capable of leveraging.
Next to the graph version, we also create a meshed variant of the
\texttt{MNIST-Superpixel} data set, first assigning to each pixel
a coordinate in $\mathbb{R}^2$ by regularly sampling the unit square,
then setting  the vertices in the simplicial complex to be the non-zero pixel
coordinates.
We then add edges and faces by computing a \emph{Delaunay complex} of
the data~(the radius of said complex spans the non-zero pixels).
The resulting complex captures both the geometry and the topology of the
images in the data set.
Following this, we classify the data using \modelname and other methods,
using a CNN architecture for the original data set and an MLP
architecture for its meshed version.
Notably, we find that our method only requires about~$20$ epochs
for training, after which training is stopped automatically, whereas
others methods use more of the allocated training budget of~$100$
epochs.
\Cref{tab:MNIST} depicts the results; we find that \modelname overall
exhibits favourable performance given its smaller footprint.
Moreover, \modelname exhibits performance on a par with competitor
methods on the meshed variant of the data set since the presence of 
higher-order structural elements like faces enables it to leverage
geometric properties.
Finally, we want to point out computational considerations.
The last column of the table shows the runtimes per epoch. Here,
\modelname outperforms all other approaches by an order of magnitude or
more.
The reported runtime is the in fact slowest of all our experiments, with
most other data sets only taking about a minute for a \emph{full}
$100$ epochs.
We report the values from \citet{Dwivedi23a}, noting that the survey uses 
a single Nvidia 1080Ti~(11GB) GPU was used on a cluster, whereas
our model was trained on a Nvidia GeForce RTX 3070 Ti~(8GB) GPU on
a commodity laptop.
This underlines the utility of \modelname as a fast,  efficient
classification method.

\begin{table}[tbp]
    \centering
    \caption{%
      Results of $5$ runs on small graph benchmark data sets.
      Parameter numbers are approximate because the number of classes differs.
      The high consistency and performance of our method on the
      `Letter' data sets is notable.
    }
    \label{tab:Benchmark datasets small}
    \sisetup{
      detect-all              = true,
      table-format            = 2.1(1),
      separate-uncertainty    = true,
      retain-zero-uncertainty = true,
      round-mode              = uncertainty,
      round-precision         = 2,
    }%
    \let\b\bfseries
    \resizebox{\linewidth}{!}{%
      \begin{tabular}{lrSSSSSS}
        \toprule
                       & {Params.} & {BZR}          &   {COX2}       & {DHFR}         & {Letter-low}   & {Letter-med}   & {Letter-high}  \\
        \midrule
        GAT            &  5K &   80.25\pm1.95 &   79.23\pm2.58 &   72.76\pm3.20 &   90.04\pm2.23 &   63.69\pm5.97  &   43.73\pm 4.13\\
        GCN            &  5K &   80.49\pm2.37 & \b79.44\pm1.78 & \b76.73\pm3.83 &   81.38\pm1.57 &   62.00\pm2.07  &   43.06\pm 1.67\\
        GIN            &  9K &   81.73\pm4.89 &   77.94\pm2.41 &   64.69\pm8.26 &   85.0 \pm0.6  &   67.07\pm2.47  &   50.93\pm 3.47\\
        \midrule
        ECT+CNN (ours) &  4K & \b81.79\pm3.19 &   70.4 \pm0.9  &   67.93\pm 4.97& \b91.51\pm2.08 & \b76.22\pm 4.76 & \b63.82\pm6.04\\
        ECT+CNN (ours) & 65K &   84.30\pm6.10 &   74.60\pm4.50 &   72.90\pm 1.60&   96.80\pm1.20 &   86.30\pm 2.00 &   85.40\pm1.30\\
        \bottomrule
      \end{tabular}%
    }%
  \end{table}

We also use a version of \modelname to classify point clouds.
In contrast to prior work~\citep{Turner14a}, we do not
use~(simplicial) complexes, but restrict the ECT to \emph{hyperplanes},
thus merely counting the number of points above or below
a given plane.
We then classify shapes from \texttt{ModelNet40} using $5$ runs,
sampling either~$100$ or~$1000$ points.
In the former case, we achieve an accuracy of $74\pm0.5$,
while in the latter case our accuracy is $77.1\pm0.4$.
Given the low complexity and high speed of our model, this is a
notable result compared to the performance reported by
\citet{zaheer2017deep}, i.e.\ $82.0\pm2.0$ and $87.0\pm2.0$,
respectively.
Moreover, \modelname is not restricted to point clouds of a specific
size, and we believe that the performance gap could potentially be
closed for models with more pronounced topological features and varying
cardinalities.

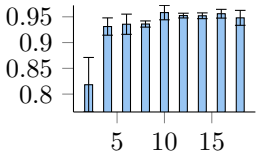
\begin{wrapfigure}[12]{l}{0.35\linewidth}
  \centering
  \begin{tikzpicture}
    \begin{axis}[%
      axis x line*          = bottom,
      axis y line*          = left,
      enlargelimits         = false,
      enlarge x limits      = true,
      error bars/y dir      = both,
      error bars/y explicit = true,
      tick align            = outside,
      ybar,
      height    = 3cm,
      width     = 4cm,
      bar width = 3.0pt,
    ]
      \addplot[fill = bleu!50, draw = black] table[%
        col sep = comma,
        x       = n_directions,
        y       = accuracy,
        y error = sdev,
      ] {results/ablation.csv};
    \end{axis}
  \end{tikzpicture}
  \caption{%
    Accuracy on the Letter-low dataset as a function of the number of
    directions.
  }
  \label{fig:Letter-low ablation}
\end{wrapfigure}
As a final experiment, we show the performance of our \modelname when it
comes to analysing graphs that contain node coordinates.
We use several graph benchmark data sets~\citep{morris2020tudataset},
with \cref{tab:Benchmark datasets small} depicting the results.
Overall, we observe high predictive performance, with \modelname
outperforming existing graph neural networks while requiring a lower
parameter budget.
We also show the benefits of substantially increasing the capacity of
our model: going to a higher parameter budget yields direct improvements
in terms of predictive performance.
Interestingly, we observe the highest gains on the `Letter' data sets,
which are subjected to increasingly larger levels of noise.
The high performance of our model in this context may point towards
better robustness properties, which we aim to investigate in future work.
Finally, as \cref{fig:Letter-low ablation} demonstrates, accuracy
remains high even when choosing a smaller number of directions for the
calculation of the ECT.

\section{Conclusion and Discussion} 
\label{sec:Conclusion and Discussion}

We described \modelname, the first differentiable framework for \emph{Euler
Characteristic Transforms}~(ECTs) and showed how to integrate it into deep learning
models.
Our method is applicable to different data modalities---including point clouds,
graphs, and meshes---and we showed its utility in a variety of learning
tasks, including both \emph{optimisation} and \emph{classification}.
The primary strength of our method is its \emph{flexibility}; it can
handle data sets with mixed modalities, containing objects with varying
sizes and shapes---we find that few algorithms such adaptability.
Moreover, our computation lends itself to high scalability and built-in
GPU acceleration; as a result, our ECT-based methods train an order of
magnitude faster than existing models on the same hardware.
We observe that our method exhibits scalability properties that
surpass existing \emph{topological machine learning}
algorithms~\citep{hensel2021survey, hajij2023topological, papamarkou2024position}.
Thus, being fully differentiable, both with respect to the number of
directions used for its calculation as well as with respect to the input
coordinates of a data set, we extend ECTs to hitherto-unavailable
applications.

\paragraph{Future work.}
We believe that this work paves the path towards new future research
directions and variants of the ECT.
Along these lines, we first aim to extend this framework to encompass
variants like the \emph{Weighted Euler Characteristic
Transform}~\citep{jiang2020weighted} or the
\emph{Smooth Euler Characteristic Transform}~\citep{Crawford20a}.
Second, while our experiments already allude to the use of the ECT to
solve inverse problems for point clouds, we would like to analyse to
what extent our framework can be used to reconstruct graphs, meshes, or
higher-order complexes.
Given the recent interest in such techniques due to their characteristic
geometrical and topological properties~\citep{Outdot20a}, we believe
that this will constitute a intriguing research direction.
Moreover, from the perspective of machine learning, there are numerous
improvements possible.
For instance, the ECT in its current form is inherently
\emph{equivariant} with respect to rotations; finding better
classification algorithms that respect this structure would thus be of
great interest, potentially leveraging spherical CNNs for improved
classification~\citep{Cohen18a}.
Our experiments on geometric graphs point towards the utility of general
geometrical-topological descriptors that offer a complementary approach
to established models based on message passing.
Leveraging existing approaches based on differential
forms~\citep{maggs2024simplicial}, we plan on establishing the ECT and
its variants as new interpretable methods for general graph learning tasks.
Finally, we aim to improve the representational capabilities of the ECT
by extending it to address node-level tasks; in this context,
topology-based methods have already exhibited favourable predictive
performance at the price of limited scalability~\citep{Horn22a}.
We hope that extensions of \modelname may serve to alleviate these issues in
the future.


\section*{Reproducibility Statement} 
\label{sec:Reproducibility Statement}

The code and configurations are provided for our experiments for reproducibility 
purposes. 
All experiments were run on a single GPU to prohibit further randomness and 
all parameters were logged.
Our code will be released under a BSD-3-Clause Licence and can be accessed under 
\href{https://github.com/aidos-lab/DECT}{\tt https://github.com/aidos-lab/DECT}.

\subsubsection*{Acknowledgments}

The authors are grateful for helpful comments provided by Jeremy
Wayland. They also wish to extend their thanks to the anonymous
reviewers and the area chair, who believed in the merits of our work.
B.R.\ is supported by the Bavarian state government with
funds from the \emph{Hightech Agenda Bavaria}.

\bibliographystyle{iclr2024_conference}
\bibliography{bibliography}

\end{document}